\documentclass[conference]{IEEEtran}
\IEEEoverridecommandlockouts

\usepackage{cite}
\usepackage{amsmath,amssymb,amsfonts}
\usepackage{graphicx}
\usepackage{xcolor}
\usepackage{url}
\usepackage{hyperref}
\usepackage{booktabs}
\usepackage{multirow}
\usepackage{tikz}
\usepackage{comment}
\usepackage{caption}
\usepackage{subcaption}
\usepackage{threeparttable}
\usepackage{xcolor}
\usepackage{siunitx}

\begin{document}

\title{Bayesian Optimization for Learning Nonlinear \\ 
        MPC in Autonomous Agent Navigation} 

\author{
    \IEEEauthorblockN{Lorenzo Ortolani, Gabriel Voss, Gabriele Beltrami, Francesco Dorati, Tommaso Felice Banfi \thanks{Corresponding author: T.~F.~Banfi, tommasofelice.banfi@talosrobotics.ai
    GitHub repository: \url{https://github.com/talos-robotics-ai/Go2_navigation}}
    }
    \IEEEauthorblockA{\textit{Talos Robotics AI} \\
	Milan, Italy \\
	\{lorenzo.ortolani, gabriel.voss, gabriele.beltrami, francesco.dorati, tommasofelice.banfi\}@talosrobotics.ai}
}

\maketitle

\begin{abstract}
Real-time autonomous navigation in dynamic, unknown environments remains a fundamental challenge for mobile robotics. We propose a map-free framework that tightly integrates reactive rolling-horizon planning with nonlinear Model Predictive Control (MPC). At each control cycle, a LiDAR-based Gaussian occupancy representation is constructed and used to generate collision-free trajectories via A* search, which are then tracked by a CasADi/IPOPT MPC formulation incorporating a smooth sigmoid obstacle barrier.
To improve robustness to parameter sensitivity, we adopt an offline Bayesian optimization scheme based on Tree-structured Parzen Estimators (TPE), which identifies near-optimal controller parameters with respect to a composite navigation objective. In addition, a Gaussian Process surrogate is used to analyze parameter sensitivity and provide insight into the optimization landscape.

The proposed framework is robot-agnostic and is evaluated on the Unitree Go2 quadruped in simulation using Gazebo, followed by deployment on the physical robot. Experimental results show that parameters tuned in simulation transfer effectively to hardware, maintaining comparable performance without additional tuning.
The full system achieves up to a 90.0\% navigation success rate when deployed, along with a 38.9\% average improvement in the evaluation metrics across simulated environments.
\end{abstract}

\begin{IEEEkeywords}
	Autonomous Agent Navigation, Model Predictive Control, Non-Linear Optimization, Bayesian Optimization
\end{IEEEkeywords}

\section{Introduction}
\label{sec:introduction}

Autonomous navigation in unknown, cluttered environments is crucial for mobile robots deployed in real-world settings. The challenge is compounded for legged platforms, where complex contact dynamics demand smooth and dynamically consistent motion commands. Classical decoupled architectures, where a global planner and low-level controller operate independently, often fail to capture the coupling between planning and dynamic feasibility~\cite{9636371}, leading to jerky execution, constraint violations, and poor out-of-sample generalization.

\begin{figure}[!ht]
	\centering

	\includegraphics[width=0.35\columnwidth]{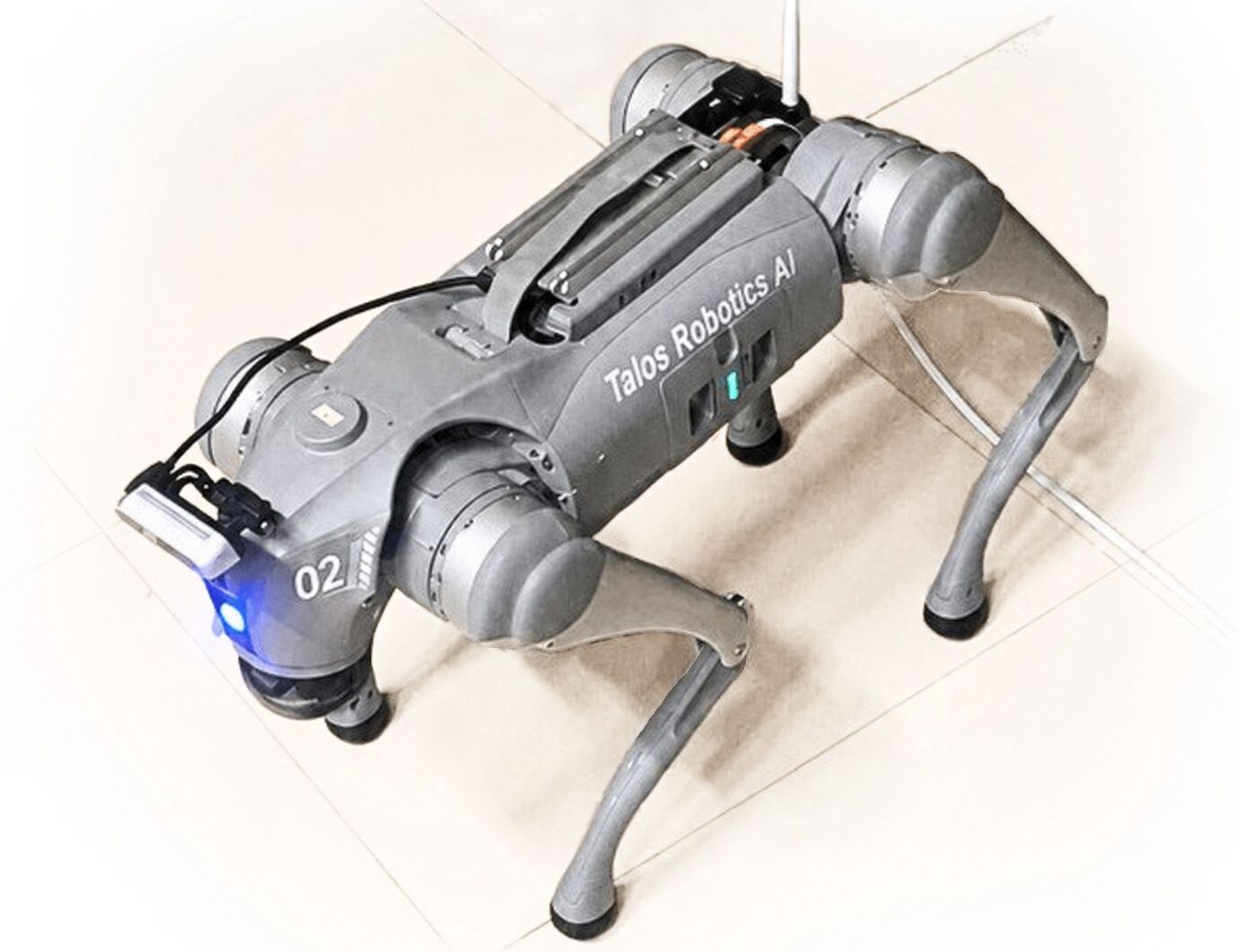}

    \caption{A customized Unitree Go2 quadruped robot by Talos Robotics AI used for our real-world navigation experiments. 
    }
	\label{fig:real-robot}
\end{figure}

Model Predictive Control (MPC) provides a principled framework for simultaneous path following and constraint satisfaction over a receding horizon~\cite{Nascimento_Dorea_Goncalves_2018}. However, real-time deployment is historically limited by two requirements: (i) a computationally efficient environment representation for obstacle avoidance, and (ii) well-calibrated cost-function weights that balance tracking accuracy, control effort, and safety. Manual tuning of these weights is time-consuming and environment-specific, with parameters that generalize poorly across different environments and navigation tasks~\cite{10924398}.

We address these limitations with a hierarchical, modular navigation stack designed as a general-purpose framework and validated on the Unitree Go2 quadruped. 
The primary contributions of this work include a map-free, rolling-horizon navigation stack combining Gaussian occupancy mapping, navigation graphs, A* planning, and nonlinear MPC with differentiable obstacle avoidance. To optimize this architecture, we introduce a Bayesian Optimization framework that uses TPE to sweep the 11-dimensional MPC parameter space across a training suite of simulation environments, identifying near-optimal weight configurations through a structured multi-metric composite score.
Finally, we provide extensive experimental validation across three environments and hardware deployment on the Unitree Go2 in Fig.~\ref{fig:real-robot} performing several real-world navigation trials, confirming consistent performance gains and negligible sim-to-real degradation.


The remainder of this paper is organized as follows.  Section~\ref{sec:background} formulates the problem and reviews related work. Section~\ref{sec:methodology} details the overall system architecture and the specifics about each module of the framework from the rolling-horizon path planner to the definition of the Bayesian Optimization pipeline. Sections~\ref{sec:experiments} and~\ref{sec:results} present the experimental setup and results, discussing the performance and limitations. Section~\ref{sec:conclusion} concludes and outlines future research directions.

\section{Background}
\label{sec:background}

\subsection{Problem Statement}
\label{sec:problem}

Let the robot state at time step $k$ be $\mathbf{x}_k = [p_{x,k},\, p_{y,k},\, \psi_k]^\top \in \mathcal{X} \subset \mathbb{R}^3$, where $(p_x, p_y) \in \mathbb{R}^2$ denotes the position in the world frame and $\psi \in (-\pi, \pi]$ is the heading angle. The control input is the body-frame velocity command $\mathbf{u}_k = [v_{x,k},\, v_{y,k},\, \omega_k]^\top \in \mathcal{U} \subset \mathbb{R}^3$, comprising forward velocity, lateral velocity, and yaw rate, respectively. At each time step the agent receives a LiDAR point cloud $\mathcal{P}_k = \{\mathbf{p}_i \in \mathbb{R}^3\}_{i=1}^{N_k}$ and a goal pose $\mathbf{x}_g \in \mathcal{X}$.

The navigation problem is formulated as a receding-horizon optimal control problem. At each planning cycle the agent solves:
\begin{equation}
	\min_{\mathbf{u}_0, \ldots, \mathbf{u}_{N-1}} \; J\!\left(\mathbf{x}_{0:N},\, \mathbf{u}_{0:N-1},\, \mathcal{P}_k,\, \theta\right)
	\label{eq:ocp}
\end{equation}
subject to:
\begin{align}
	\mathbf{x}_{t+1}               & = f(\mathbf{x}_t, \mathbf{u}_t), \quad t = 0, \ldots, N{-}1 \label{eq:dyn} \\
	d(\mathbf{x}_t, \mathcal{P}_k) & \geq r_{\min}, \quad t = 0, \ldots, N \label{eq:obs}                       \\
	\mathbf{u}_t                   & \in \mathcal{U}, \quad t = 0, \ldots, N{-}1 \label{eq:ctrl}
\end{align}
where $J$ is a composite cost function parametrized by $\theta \in \Theta \subset \mathbb{R}^{n_\theta}$, $f$ is the discrete-time robot dynamics, $d(\mathbf{x}_t, \mathcal{P}_k)$ is the minimum Euclidean distance from the predicted state $\mathbf{x}_t$ to any obstacle in the current point cloud, and $r_{\min}$ is the minimum safety clearance. The objective is to drive the robot from its current pose $\mathbf{x}_0$ to the goal $\mathbf{x}_g$ while satisfying \eqref{eq:dyn}--\eqref{eq:ctrl} in real time, without any a priori knowledge of the environment.

The performance of the MPC \eqref{eq:ocp} depends critically on the parameter vector $\theta$, which encodes cost weights for tracking accuracy, control effort, and obstacle avoidance. The meta-problem of identifying an optimal $\theta$ is formulated as the black-box optimization:
\begin{equation}
	\theta^* = \arg\max_{\theta \in \Theta} \; \mathcal{J}(\theta)
	\label{eq:meta}
\end{equation}
where $\mathcal{J}(\theta)$ is a composite performance score obtained by running the full navigation system under parameters $\theta$.

\subsection{Related Work}

\subsubsection{Legged Robot Navigation}
Navigation for legged robots has been studied extensively in the context of whole-body planning and terrain-aware locomotion. Recent approaches integrate deep learning-based terrain estimation with classical planning~\cite{8392399} or employ model-free reinforcement learning for end-to-end navigation~\cite{doi:10.1126/scirobotics.aau5872}. These methods either require pre-built maps or suffer from difficult sim-to-real transfer. 
Our framework abstracts leg dynamics through the CHAMP hierarchical locomotion controller~\cite{articleHierarchicalController}, which exposes a body-frame velocity interface and enables the navigation layer to treat the robot as a holonomic mobile agent, simplifying planning and control design.

For local navigation, fast sampling-based planners like RRT*~\cite{karaman2011sampling} and adaptive variants such as APEI-RRT*~\cite{10796871} are highly efficient and widely adopted. While we deploy a grid-based A* planner for its predictable deterministic paths, our framework is fundamentally planner-agnostic; the adaptive MPC pipeline can seamlessly track reference trajectories generated by any sampling-based or heuristic algorithm.

\subsubsection{MPC for Mobile Robot Navigation}
MPC has been applied to mobile robot navigation across differential-drive platforms, UAVs~\cite{10896285}, and quadrupeds. Nonlinear MPC formulations handle the full kinematic model but require efficient nonlinear programming (NLP) solvers; CasADi~\cite{andersson2018casadi} with IPOPT~\cite{wachter2006implementation} provides a suitable combination for our 5-second horizon at 10\,Hz. Obstacle avoidance within MPC is commonly implemented via hard distance constraints or smooth barrier functions~\cite{ames2019control}; we adopt a differentiable sigmoid penalty evaluated on the nearest LiDAR returns, which preserves NLP sparsity and IPOPT convergence properties.

\subsubsection{Bayesian Optimization for Robotics}
Bayesian Optimization (BO) has become the standard method for black-box hyperparameter tuning in robotics, widely applied to locomotion, control, and planning~\cite{calandra2016bayesian}. The COAt-MPC framework~\cite{10924398} provides performance-guaranteed BO for MPC parameter optimization, while~\cite{watanabe2025treestructuredparzenestimatorunderstanding} analyzes the theoretical properties of TPE, the acquisition strategy we adopt for its efficiency in high-dimensional mixed spaces. 

\section{Methodology}
\label{sec:methodology}

\subsection{System Architecture}

\begin{figure}[!ht]
	\centering
	\includegraphics[width=0.77\columnwidth]{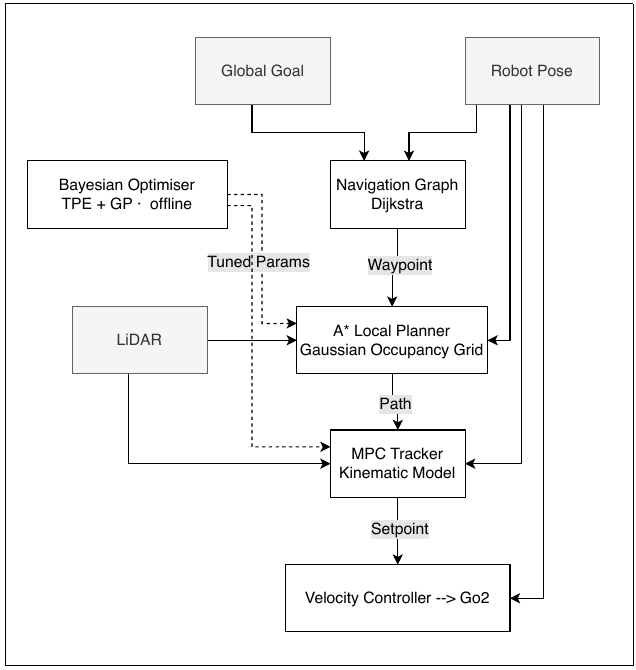}
\caption{System architecture of the proposed framework.}	
\label{fig:system-architecture}
\end{figure}

The proposed navigation stack is composed of five interconnected modules illustrated in Fig.~\ref{fig:system-architecture}: (i) an \textit{Odometry Bridge} that converts raw EKF odometry into a unified pose representation; (ii) a \textit{Mapping Module} that maintains both a reactive semi-persistent Gaussian occupancy grid and a topological \textit{Navigation Graph} inspired by WildOS~\cite{shah2026wildosopenvocabularyobjectsearch}, which encodes high-level connectivity between previously visited locations and enables global planning via Dijkstra search; (iii) a \textit{Global-to-Local Planning Layer}, where the Navigation Graph provides waypoints to a rolling-horizon A* planner operating on the occupancy grid at 1\,Hz; (iv) a \textit{Nonlinear MPC Tracker} formulated with CasADi~\cite{andersson2018casadi} and solved with IPOPT~\cite{wachter2006implementation}, generating dynamically feasible velocity setpoints and converting MPC lookahead states into body-frame \texttt{/cmd\_vel} commands for the locomotion layer; and (v) an offline \textit{Bayesian Optimisation Module} based on Tree-structured Parzen Estimator (TPE) and Gaussian Processes (GP), which tunes the hyperparameters to improve trajectory efficiency and tracking performance.

All modules communicate through ROS~2 topics, enabling full modularity and real-time operation at 10--20\,Hz. The MPC parameter vector $\theta$ is adapted by the Bayesian Optimization pipeline in Section~\ref{sec:offline-bo}.

\subsection{Mapping}

At each replanning cycle, a square local Gaussian Occupancy Grid of side $2h_w$ (default $h_w = 5\,\text{m}$, resolution $\Delta = 0.25\,\text{m/cell}$) is centered on the robot and rebuilt from scratch using only the current LiDAR point cloud $\mathcal{P}_k$. Each grid cell $c$ encodes the probability of occupancy derived from the minimum Euclidean distance $d_{\min}(c, \mathcal{P}_k)$ from the cell center to any LiDAR return:
\begin{equation}
	P(c) = 1 - \Phi\!\left(\frac{d_{\min}(c,\, \mathcal{P}_k)}{\sigma}\right)
	\label{eq:occ}
\end{equation}
where $\Phi(\cdot)$ is the standard normal CDF and $\sigma = 0.05\,\text{m}$ is the Gaussian spread parameter controlling obstacle inflation. Cells with $P(c) \geq P_{\text{thresh}} = 0.45$ are treated as hard obstacles; cells below this threshold incur a soft traversal penalty in A* proportional to their occupancy probability.

In addition to the reactive occupancy grid, the system maintains a persistent topological navigation graph $\mathcal{G} = (\mathcal{V}, \mathcal{E})$, where each node $v \in \mathcal{V}$ corresponds to a semantically meaningful region of the environment and each edge $(u,v) \in \mathcal{E}$ represents a navigable transition weighted by the traversal cost estimated from the current occupancy grid. As the robot explores, new nodes are instantiated when the robot enters a previously unvisited region, and the graph is updated online.

\subsection{Rolling-Horizon Path Planning}

Since the occupancy grid covers only a local window, a rolling-horizon strategy is employed for distant goals. At each replanning tick the local planning target is:
\begin{equation}
	\mathbf{x}_{\text{local}} = \begin{cases}
		\text{cell}(\mathbf{x}_g)                                          & \text{if } \mathbf{x}_g \in \text{grid} \\
		\text{ray}(\mathbf{x}_0 \to \mathbf{x}_g) \cap \partial\text{grid} & \text{otherwise}
	\end{cases}
	\label{eq:local-goal}
\end{equation}
The ray is parametrically clipped to the grid boundary, allowing the robot to advance one grid-width at a time. If the local target falls in an occupied cell, a breadth-first search (BFS) identifies the nearest free cell as a proxy, guaranteeing A* always receives a valid goal.

A* is executed on an 8-connected grid with the following edge traversal cost:
\begin{equation}
	g(n \to n') = c_{\text{move}} \cdot \Delta \cdot \left(1 + w_{\text{obs}} \cdot P(n')\right)
	\label{eq:astar-cost}
\end{equation}
where $c_{\text{move}} \in \{1, \sqrt{2}\}$ for cardinal and diagonal moves respectively and $w_{\text{obs}}$ is the soft obstacle penalty weight. The admissible Euclidean heuristic is:
\begin{equation}
	h(n) = \Delta \cdot \sqrt{(i_x - g_x)^2 + (i_y - g_y)^2}
	\label{eq:heuristic}
\end{equation}
Cells with $P \geq P_{\text{thresh}}$ are treated as hard obstacles, while sub-threshold cells incur the soft penalty~\eqref{eq:astar-cost}, steering paths away from obstacle proximity. The resulting path is resampled at uniform spacing $\Delta s = 0.20\,\text{m}$ and smoothed with a moving-average kernel of width 5 to remove grid-induced zigzag artifacts while preserving endpoints.

\subsection{MPC Formulation}
\label{sec:mpc}

\subsubsection{State, Control, and Dynamics}

The robot is modeled as a holonomic kinematic agent in the plane, with state $\mathbf{x}_k = [p_x, p_y, \psi]^\top$ and control $\mathbf{u}_k = [v_x, v_y, \omega]^\top$. The discrete-time dynamics under Forward Euler integration with step $\Delta t = 0.1\,\text{s}$ are:
\begin{equation}
	\mathbf{x}_{k+1} = f(\mathbf{x}_k, \mathbf{u}_k) = \begin{bmatrix}
		p_{x,k} + (v_{x,k}\cos\psi_k - v_{y,k}\sin\psi_k)\,\Delta t \\
		p_{y,k} + (v_{x,k}\sin\psi_k + v_{y,k}\cos\psi_k)\,\Delta t \\
		\psi_k + \omega_k\,\Delta t
	\end{bmatrix}
	\label{eq:dynamics}
\end{equation}

\subsubsection{Objective Function}

The MPC solves the following finite-horizon NLP over $N = 50$ steps (5\,s horizon) at 10\,Hz:
\begin{align}
	J =\; & \mathbf{e}_N^\top \mathbf{Q}_T \mathbf{e}_N + J_{\text{obs}}(\mathbf{x}_N) \nonumber           
	       + \sum_{k=0}^{N-1} \Bigl[ \mathbf{e}_k^\top \mathbf{Q} \mathbf{e}_k \\
	       & + \mathbf{u}_k^\top \mathbf{R} \mathbf{u}_k + R_{\text{jerk}} \|\mathbf{u}_k - \mathbf{u}_{k-1}\|^2 + J_{\text{obs}}(\mathbf{x}_k) \Bigr]
	\label{eq:mpc-cost}
\end{align}
where $\mathbf{e}_k = \mathbf{x}_k - \mathbf{x}_{\text{ref},k}$ is the tracking error with respect to the A*-derived reference, $\mathbf{Q} = \text{diag}(Q_{xy}, Q_{xy}, Q_\psi)$, $\mathbf{Q}_T = Q_T \cdot \mathbf{Q}$, and $\mathbf{R} = \text{diag}(R_v, R_v, R_\omega)$. The jerk term $R_{\text{jerk}}$ penalizes rapid control changes to enforce smooth commands. The NLP is compiled once at startup using CasADi~\cite{andersson2018casadi} symbolic differentiation; subsequent solves update only numeric parameter values, enabling real-time feasibility. 
The MPC predicted trajectory is tracked by a proportional controller running at 20\,Hz, which generates \texttt{/cmd\_vel} commands for the CHAMP locomotion layer.

\subsubsection{Sigmoid Obstacle Barrier}

Obstacle avoidance is enforced as a smooth cost penalty through a logistic sigmoid barrier evaluated on the $M = 12$ nearest LiDAR returns $\{\mathbf{p}_j\}_{j=1}^M$ within a search radius of 3\,m:
\begin{equation}
	J_{\text{obs}}(\mathbf{x}_k) = \sum_{j=1}^{M} \frac{W}{1 + e^{\alpha\bigl(d(\mathbf{x}_k, \mathbf{p}_j) - r\bigr)}}
	\label{eq:barrier}
\end{equation}
where $d(\mathbf{x}_k, \mathbf{p}_j) = \sqrt{(p_{x,k} - p_{j,x})^2 + (p_{y,k} - p_{j,y})^2 + \epsilon}$ with $\epsilon = 10^{-6}$ ensuring differentiability at zero distance. The numerically stable form $\frac{W}{2}\bigl(1 - \tanh\!\bigl(\frac{\alpha}{2}(d - r)\bigr)\bigr)$ is used in implementation to avoid overflow. Points beyond the search radius are replaced with far sentinels, keeping the NLP sparsity pattern constant and enabling warm starting. Here $W$ is the barrier height, $\alpha$ controls steepness, and $r$ is the safety radius.
The control inputs are enforced via box constraints corresponding to the robot actuator limits.

\subsubsection{Reference Trajectory and Warm Starting}

The reference sequence $\{\mathbf{x}_{\text{ref},k}\}_{k=0}^N$ is generated by advancing along the A* path at cruise speed $v_{\text{ref}}$ from the arc-length coordinate $s_0$ of the closest waypoint to the robot:
\begin{equation}
	s_k = \min\!\left(s_0 + v_{\text{ref}} \cdot k \cdot \Delta t,\; L_{\text{path}}\right)
	\label{eq:ref-traj}
\end{equation}
Reference position $\mathbf{p}_{\text{ref},k}$ is linearly interpolated along the path segment containing $s_k$; reference yaw is the path tangent $\psi_{\text{ref},k} = \text{atan2}(\Delta y, \Delta x)$. The previous IPOPT solution is shifted by one step and used as the warm-start initial guess. 

\subsection{Bayesian Optimization for MPC Tuning}
\label{sec:offline-bo}

\subsubsection{Composite Performance Score}

The MPC parameter vector $\theta \in \Theta \subset \mathbb{R}^{11}$ comprises: position tracking weights $(Q_x, Q_y)$, yaw tracking weight $Q_{\psi}$, terminal cost weight $Q_T$, control effort weights $(R_{v_x}, R_{v_y}, R_{\omega})$, jerk regularization weight $R_{\text{jerk}}$, obstacle cost weight $W_{\text{obs}}$, and obstacle shaping parameters $(\alpha_{\text{obs}}, r_{\text{obs}})$.
Each parameter is associated with predefined lower and upper bounds, defining the search space as a bounded domain $\Theta = \prod_i [\theta_i^{\min}, \theta_i^{\max}]$.

Each function evaluation requires launching a full Gazebo simulation, making gradient-free, sample-efficient methods essential. The composite score $\mathcal{J}(\theta)$ aggregates different navigation scenarios $s \in S$ with scenario weights $w_s$:
\begin{equation}
	\mathcal{J}(\theta) = \sum_{s} w_s \cdot \mathcal{J}_s(\theta), \qquad \sum_s w_s = 1
	\label{eq:agg-score}
\end{equation}

A scenario includes various types of trajectories imposed on the robot in simulation, enabling it to learn optimal parameters under diverse dynamic configurations.
Each per-scenario score $\mathcal{J}_s$ integrates five metrics. Let $\mathbf{p}_0$, $\mathbf{p}_f$, and $\mathbf{g}$ denote the robot start, final, and goal positions. \textit{Goal progress} $\phi = \max(0, (d_0 - d_f)/d_0)$ measures the fraction of initial distance closed, where $d_0 = \|\mathbf{p}_0 - \mathbf{g}\|$, $d_f = \|\mathbf{p}_f - \mathbf{g}\|$. \textit{Path efficiency} $\eta = \min(1, d_0/L)$ is the ratio of the straight-line distance to the total traversed arc length $L$. \textit{Control smoothness} $s = e^{-\bar{j}/2}$ uses the exponentiated negative mean second-order finite difference of the command sequence as a jerk proxy:
\begin{equation}
	\bar{j} = \frac{1}{K-2}\sum_{k=1}^{K-2} |\Delta^2 \mathbf{u}_k|_1, \quad \Delta^2 \mathbf{u}_k = \mathbf{u}_{k+1} - 2\mathbf{u}_k + \mathbf{u}_{k-1}
	\label{eq:jerk}
\end{equation}
\textit{Obstacle avoidance} $\mathcal{O}$ uses per-scan minimum LiDAR distances $\{d_k^{\text{obs}}\}$ to define danger ($d < 0.3\,\text{m}$) and warning ($0.3 \le d < 0.6\,\text{m}$) zone fractions:
\begin{equation}
	\mathcal{O} = 0.5\,(1 - f_{\text{danger}}) + 0.3\,(1 - f_{\text{warning}}) + 0.2\,\min\!\left(\tfrac{\bar{d}}{2}, 1\right)
	\label{eq:obs-score}
\end{equation}
where $\bar{d}$ is the mean clearance capped at 2\,m. \textit{Time efficiency} $\tau = \min(1, T_{\text{ref}}/T_{\text{elapsed}})$ with $T_{\text{ref}} = d_0/0.5$ is only defined when the goal is reached.
Two scoring branches are defined:
\begin{equation}
	\mathcal{J}_s = \begin{cases}
		0.35 + 0.2\,\eta + 0.15\,s + 0.2\,\mathcal{O} + 0.1\,\tau & \text{if goal reached}     \\
		0.25\,\phi + 0.1\,\eta + 0.1\,s + 0.15\,\mathcal{O}        & \text{else}
	\end{cases}
	\label{eq:score-branch}
\end{equation}

\subsubsection{Tree-Structured Parzen Estimator}

The offline BO employs Tree-Structured Parzen Estimators (TPE)~\cite{watanabe2025treestructuredparzenestimatorunderstanding}, well-suited for high-dimensional mixed parameter spaces. TPE partitions the observed scores at the $\gamma = 0.15$ quantile threshold $y^*$ and models two marginal densities:
\begin{equation}
	\ell(\theta) = p(\theta \mid \mathcal{J}(\theta) < y^*), \quad g(\theta) = p(\theta \mid \mathcal{J}(\theta) \geq y^*)
	\label{eq:tpe-densities}
\end{equation}
Each marginal is estimated as a mixture of truncated Gaussians, one centred on each past observation. The acquisition is the Expected Improvement ratio:
\begin{equation}
	\text{EI}(\theta) \propto \frac{\ell(\theta)}{g(\theta)}
	\label{eq:tpe-acq}
\end{equation}

The algorithm first draws $N_{\text{rand}} = 20$ trials uniformly (cold start), then iterates: refit $\ell$ and $g$, sample 20 candidates from $\ell(\theta)$, select the one with highest $\ell/g$ ratio, evaluate $\mathcal{J}$, and add to the dataset. This repeats until $N_{\text{max}} = 120$ trials and the best configuration $\theta^*_{\text{offline}} = \arg\max_{i} \mathcal{J}(\theta_i)$ is deployed.

\subsubsection{GP Surrogate for Sensitivity Analysis}

An ARD Matérn-5/2 Gaussian Process is fitted to all accumulated observations for interpretability. The kernel is:
\begin{equation}
	k(\theta, \theta') = \sigma_f^2 \!\left(1 + \sqrt{5}\,r + \tfrac{5}{3}r^2\right)\!e^{-\sqrt{5}\,r} + \sigma_n^2 \delta(\theta, \theta')
	\label{eq:gp-kernel}
\end{equation}
where $r = \sqrt{\sum_i (\theta_i - \theta_i')^2/\ell_i^2}$ and ARD length scales $\{\ell_i\}$ are learned by maximizing the log marginal likelihood. The normalized inverse length-scale $\tilde{\ell}_i = (1/\ell_i)/\sum_j(1/\ell_j)$ quantifies parameter sensitivity: a short $\ell_i$ indicates that the score landscape varies rapidly with $\theta_i$, marking it as decisive.

\section{Experimental Setup}
\label{sec:experiments}

All experiments are conducted on the Unitree Go2 quadruped. Locomotion is handled by the CHAMP hierarchical controller~\cite{articleHierarchicalController}, which exposes a body-frame \texttt{/cmd\_vel} velocity interface. A mid-range 3-D LiDAR provides filtered point clouds at 10\,Hz. The full navigation stack uses Ros~2 and runs on a Jetson Orin Nano carried onboard, while the 
simulations are performed in Gazebo Fortress~\cite{koenig2004design} with a planar-move plugin providing ground-truth odometry.

Three Gazebo worlds of increasing complexity are employed. Environments E1 and E2 serve as in-distribution training environments for offline Bayesian Optimization; E3 is held out as an out-of-distribution test scenario.

\begin{itemize}
	\item \textbf{E1 -- Open} (tuning): a sparse, low-obstacle environment representing easy navigation conditions.
	\item \textbf{E2 -- Indoor Office} (tuning): a confined indoor environment with tables, narrow passages, doorways, and multiple rooms, presenting high navigation difficulty.
	\item \textbf{E3 -- Warehouse} (testing): a held-out environment with dense shelves, narrow corridors, and complex passages, designed to assess out-of-distribution generalization.
	\item \textbf{E4 -- Real Room} (testing): a real room of our laboratory arranged for the experiment with obstacles as furniture, boxes and moving people.
\end{itemize}

Each environment contains a representative set of navigation tasks of varying difficulty, including straight-line traversals, zig-zag paths, tight turns, navigation around large obstacles, and goal configurations requiring global exploration. 
An additional test with suddenly appearing dynamic obstacles evaluates robustness to pedestrians or vehicles in unstructured settings.
The evaluation is composed of the following phases:
\begin{itemize}
    \item \textbf{Phase~0 -- Baseline:} MPC parameters are hand-tuned through a limited set of simulation trials and held fixed, which serves as the reference for all comparisons.
    \item \textbf{Phase~1 -- Offline BO:} 120 BO trials are executed on E1 and E2 using the TPE method described in Section~\ref{sec:offline-bo}, yielding the tuned vector $\theta^*_{\text{BO}}$.
    \item \textbf{Phase~2 -- Generalization:} Both configurations are evaluated on the unseen environment E3 to quantify out-of-distribution transfer of BO tuned parameters.
    \item \textbf{Phase~3 -- Sim-to-Real:} We performed a limited test in a real environment where the robot had to reach predefined set of navigation tasks in the environment E4.
\end{itemize}
As evaluation metrics for the navigation experiments, we report success rate (SR, \%), time to goal (TTG, s), path length (PL, m), and average MPC solve time (TSM, ms). 
Success rate (SR) is defined as the percentage of episodes in which the robot reaches the goal region within a predefined tolerance, without collision, and within a 5-minute time limit. Time to goal (TTG) measures the total elapsed time from the initial pose to goal attainment. Path length (PL) denotes the cumulative distance travelled along the executed trajectory. 
Finally, average MPC solve time (TSM) quantifies computational cost as the mean solver execution time per MPC iteration.

\section{Results}
\label{sec:results}

\begin{table*}[!ht]
\centering
\begin{threeparttable}
\caption{As navigation evaluation metrics, we report success rate (SR, \% or success/trials, $\uparrow$), time to goal (TTG, s, $\downarrow$), path length (PL, m, $\downarrow$) and average time to solve MPC (TSM, ms, $\uparrow$). Note: $\uparrow$ denotes higher is better, $\downarrow$ lower is better.}

\label{tab:performance}

\renewcommand{\arraystretch}{1.15}
\begin{tabular}{lccccccccccc}
\toprule
& \multicolumn{2}{c}{\textbf{E1 (Open)}} 
& \multicolumn{2}{c}{\textbf{E2 (Indoor Office)}} 
& \multicolumn{2}{c}{\textbf{E3\tnote{*} (Warehouse)}} 
& \multicolumn{4}{c}{\textbf{E\textsubscript{avg(1,2,3)} (Overall)}} 
& \textbf{E4\tnote{*} (Real Room)} \\
\cmidrule(lr){2-3}
\cmidrule(lr){4-5}
\cmidrule(lr){6-7}
\cmidrule(lr){8-11}
\cmidrule(lr){12-12}
\textbf{Method} & PL & TSM & PL & TSM & PL & TSM & PL & TSM & TTG & SR & SR \\
\midrule
Baseline parameters $\theta_0$
& 9.64 
& 40.85 
& 17.92 
& 46.78 
& 73.97 
& 60.78 
& 33.84 
& 49.47 
& 200.37 
& 65.0\%
& 5/10\\

\textbf{BO tuned parameters} $\theta^*_{\text{BO}}$
& 9.03 
& 38.68 
& 12.13 
& 23.77 
& 41.07 
& 48.41 
& 20.74 
& 36.95 
& 94.25 
& 90.0\%
& 8/10\\

Percentage change $\Delta_\%$
& \textcolor{green!40!black}{-6.3}
& \textcolor{green!45!black}{-5.3}
& \textcolor{green!45!black}{-32.3} 
& \textcolor{green!45!black}{-49.2} 
& \textcolor{green!45!black}{-44.5} 
& \textcolor{green!45!black}{-20.4} 
& \textcolor{green!45!black}{-38.7} 
& \textcolor{green!45!black}{-25.3} 
& \textcolor{green!45!black}{-53.0} 
& \textcolor{green!45!black}{+38.5}
& \textcolor{green!45!black}{+60.0} \\
\bottomrule
\end{tabular}

\begin{tablenotes}
\footnotesize
\item[*] Indicates held-out out-of-distribution environment.
\end{tablenotes}

\end{threeparttable}
\end{table*}

\subsection{Navigation Performance}

\begin{figure}[!ht]
	\centering
    \text{MPC objective cost function {\footnotesize (log-scale)}}
	\includegraphics[width=1.0\columnwidth]{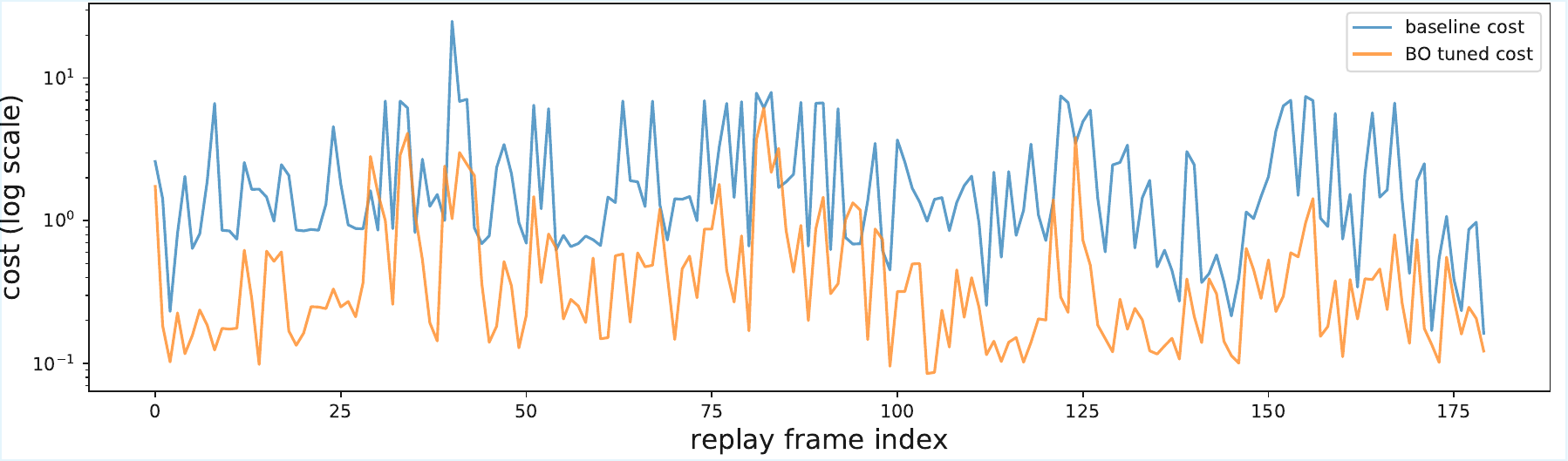}
	\caption{Comparison of the MPC objective function~\eqref{eq:mpc-cost} over the planning horizon between baseline and BO-tuned configurations in E3. Lower values indicate tighter path tracking and reduced obstacle cost, so overall better task performance.}
	\label{fig:MPC_objective_function_log}
\end{figure}

As shown in Table~\ref{tab:performance}, the hand-tuned baseline performs adequately in the open environment (E1: PL\,=\,9.64\,m, TSM\,=\,40.85\,ms) but degrades in confined settings (E2: PL\,=\,17.92\,m, TSM\,=\,46.78\,ms), reflecting the sensitivity of fixed parameters to obstacle density. Offline BO on E1 and E2 consistently improves all metrics across every evaluated scenario. In the indoor office environment (E2), path length is reduced by 32.3\% (17.92\,m\,$\to$\,12.13\,m) and average MPC solve time decreases by 49.2\% (46.78\,s\,$\to$\,23.77\,s), indicating that BO-tuned parameters yield shorter, more computationally efficient trajectories. Critically, The BO-tuned configuration generalizes to the unseen warehouse environment (E3): path length drops by 44.5\% (73.97\,m\,$\to$\,41.07\,m), despite E3 never being used during tuning, demonstrating strong out-of-distribution transfer.  
Figure~\ref{fig:MPC_objective_function_log} illustrates the MPC objective function~\eqref{eq:mpc-cost} comparison, which is consistently lower.

Aggregated over E1--E3, BO reduces average path length by 38.7\%, average MPC solve time by 25.3\%, and time to goal by 53.0\% (200.37\,s\,$\to$\,94.25\,s), while improving the overall success rate from 13/20 to 18/20. These gains confirm that optimizing across environments of contrasting difficulty produces robust, generalizing parameter configurations. 

\begin{figure}[!ht]
	\centering
	\begin{subfigure}[t]{0.58\columnwidth}
		\centering
		\vspace{-0.85em}
		\fbox{\includegraphics[height=5cm]{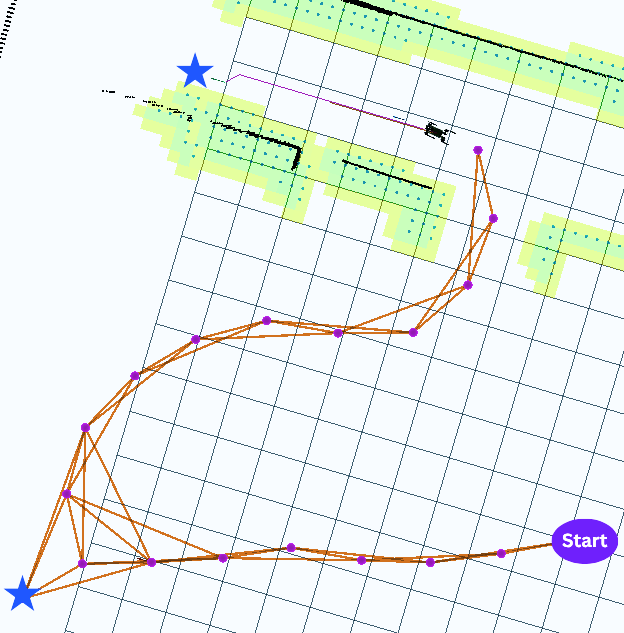}}
        \caption{Topological navigation graph constructed online during the robot’s exploration phase, where visited locations are nodes and edges encode regions connectivity. The purple circle marks the starting point, the blue stars are the first reached goal and the current target. The robotic agent is maneuvering around obstacles toward the second goal.}
        
    \end{subfigure}
	\hfill
	\begin{subfigure}[t]{0.40\columnwidth}
		\centering
		\vspace{-0.85em}
		\includegraphics[height=5.35cm]{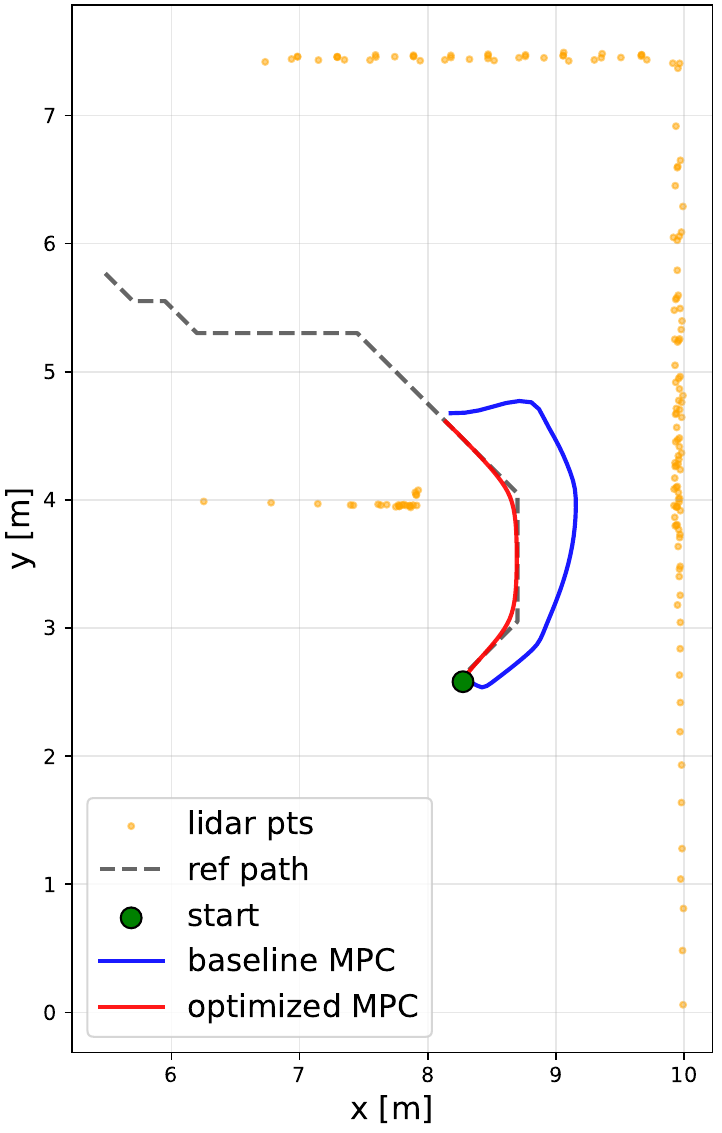}
        \caption{Trajectory comparison in E3 showing A* reference (grey dotted), baseline MPC (blue), and BO-optimized MPC (red). Current robot pose is green; LiDAR points of obstacles are shown in yellow.}
         \end{subfigure}

	\caption{Navigation graph and MPC path planner comparison in the held out-of-distribution warehouse environment (E3).}
	\label{fig:combined_plots}
\end{figure}

Figure~\ref{fig:combined_plots} illustrates a representative trajectory comparison in E3, where the BO-tuned MPC follows a shorter, smoother path relative to the baseline.

\setlength{\fboxrule}{0.3pt}
\setlength{\fboxsep}{0pt}

Hardware deployment in the real laboratory room (E4) validates sim-to-real transfer: success rate improves from 5/10 to 8/10 using $\theta^*_{\text{BO}}$ without any additional on-robot or online tuning, confirming that simulation-tuned parameters transfer effectively to physical hardware and real-world obstacles.

\subsection{MPC Parameter Sensitivity}
\begin{figure}[!ht]
	\centering
    \text{Bayesian optimization convergence}
	\includegraphics[width=0.84\columnwidth]{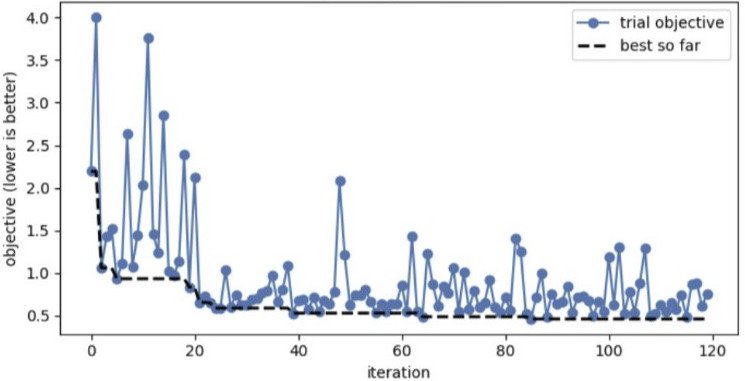}
    \caption{Bayesian optimization (BO) convergence, showing the objective score $\mathcal{J}(\theta)$ versus trial index over 120 trials. 
    }
	\label{fig:MPCconvergence}
\end{figure}

We analyze MPC parameter importance via GP-based sensitivity estimation (Section~\ref{sec:offline-bo}), using baseline and BO-tuned configurations. The BO convergence is shown in Fig.~\ref{fig:MPCconvergence}.

The highest sensitivity is observed for control jerk $R_{\text{jerk}}$ and control effort $R_v$, indicating that dynamic regularization primarily governs solution stability. Obstacle weight $W$ and solver-time effects follow, while tracking terms ($Q_{xy}$, $Q_\psi$) and failure penalty show low sensitivity.
Overall, variability is driven by dynamic and constraint-related parameters, with tracking having limited influence. Compared to the baseline, BO reduces average sensitivity by 42.3\%, yielding a smoother landscape that improves robustness and generalization.

\section{Discussion and Limitations}
\label{sec:discussion}

\textbf{Limitations.} The kinematic model treats the robot as a planar holonomic agent; significant slopes or steps require extension to 3-D dynamics
. 
The BO search is bounded by the predefined space $\Theta$, which may not include all the possible MPC parameters that could be tuned to find the optimal configuration, an a priori selection was necessary to avoid an extremely high dimensional optimization space.

\textbf{Future Work.} Three primary extensions are planned. First, an online BO loop warm-started from $\theta^*_{\text{BO}}$ will adapt MPC parameters incrementally during deployment via GP posterior updates and Expected Improvement acquisition~\cite{calandra2016bayesian}, enabling environment-specific refinement for out-of-distribution settings without offline retraining. Second, a self-supervised Dynamic Graph CNN (DGCNN) encoder will provide scan-level, annotation-free MPC parameter conditioning from LiDAR embeddings. Third, the stack will be ported to the Unitree G1 bipedal humanoid and interfaced with Visual Language Action (VLA)~\cite{pmlr-v229-zitkovich23a}
models for natural-language goal decomposition, with lifelong continual online learning 
to accumulate knowledge from the real world environment data.

\section{Conclusion}
\label{sec:conclusion}

We have presented a map-free autonomous navigation framework combining rolling-horizon A* planning with nonlinear MPC tuned via offline Bayesian Optimization. The system integrates a Gaussian occupancy grid, a CasADi/IPOPT nonlinear MPC with differentiable sigmoid obstacle barriers, and a TPE-based BO pipeline that identifies near-optimal MPC weight configurations from a composite performance score.

Offline BO conducted on two training environments of different difficulty yields consistent improvements across all evaluated scenarios. The average length path decreases by 38.7\% and the MPC average total solving time decreases by 25.2\% relative to the hand-tuned baseline. The BO-tuned configuration generalizes to the held-out warehouse environment without additional tuning.

The modular architecture is platform-agnostic and designed for extension. Our ongoing research work targets online Bayesian Optimization for deployment-time adaptation, DGCNN for map-adaptive MPC parameter conditioning, and integration with Visual Language Action models for language-conditioned navigation across legged and wheeled platforms.

\section*{Acknowledgment}
\label{sec:acknowledgment}

This work was conducted by the research team at Talos Robotics AI (\url{https://talosrobotics.ai/}). 
The authors acknowledge the primary technical contribution of L.~Ortolani to the development of the proposed framework. 
T.~F.~Banfi served as the first co-author and corresponding author; he is also affiliated with ETH Zurich and CERN; however, his contribution is not related to these institutions.

\bibliographystyle{IEEEtran}
\bibliography{bibliography}

\end{document}